\documentclass[sigconf]{acmart}
\AtBeginDocument{%
  }

\setcopyright{acmlicensed}
\copyrightyear{2018}
\acmYear{2018}
\acmDOI{XXXXXXX.XXXXXXX}

\acmConference[Conference acronym 'XX]{Make sure to enter the correct
  conference title from your rights confirmation email}{June 03--05,
  2018}{Woodstock, NY}

\acmISBN{978-1-4503-XXXX-X/2018/06}

\usepackage{multirow}
\usepackage[ruled,vlined,linesnumbered,noend]{algorithm2e}
\usepackage{listings}
\usepackage{xspace}
\usepackage{tabularx}
\usepackage{array}

\usepackage{enumitem}
\usepackage{listings}

\usepackage{fancyhdr}
\fancypagestyle{plain}{%
  \fancyhf{}%
  \fancyfoot[C]{\thepage}%
}
\lstdefinestyle{promptstyle}{%
  basicstyle=\footnotesize\ttfamily,
  breaklines=true,
  breakatwhitespace=true,
  frame=leftline,
  framesep=6pt,
  xleftmargin=10pt,
  rulesep=0pt,
  columns=fullflexible,
  keepspaces=true,
  showstringspaces=false,
  aboveskip=6pt,
  belowskip=6pt,
}
\lstnewenvironment{prompt}{\lstset{style=promptstyle}}{}
\usepackage{xspace}
\newcommand{\sysname}{AutoKD\xspace}

\usepackage[table]{xcolor}

\usepackage[most]{tcolorbox}

\newtcolorbox{promptbox}[1]{
    enhanced,
    breakable,
    colback=blue!5!white,
    colframe=blue!75!black,
    fonttitle=\mdseries,
    title=#1,
    boxrule=0.5pt,
    arc=3pt
}
\newcommand{\pvar}[1]{\texttt{\{\detokenize{#1}\}}}
\newcommand{\pcode}[1]{\texttt{\detokenize{#1}}}

\begin{document}

\title{AutoKD: Autonomous Knowledge Discovery}

\author{Qinwen Ge}
\affiliation{\institution{Vanderbilt University}
\country{}}
\email{qinwen.ge@vanderbilt.edu}

\author{Bo Ni}
\affiliation{\institution{Vanderbilt University}
\country{}}
\email{bo.ni@vanderbilt.edu}

\author{Haowei Fu}
\affiliation{\institution{Vanderbilt University}
\country{}}
\email{haowei.fu@vanderbilt.edu}

\author{Ngoc N. Tran}
\affiliation{\institution{Vanderbilt University}
\country{}}
\email{ngoc.n.tran@vanderbilt.edu}

\author{Erik Blasch}
\affiliation{\institution{Air Force Research Laboratory}
\country{}}
\email{erik.blasch.1@us.af.mil}

\author{Tyler Derr}
\affiliation{\institution{Vanderbilt University}
\country{}}
\email{tyler.derr@vanderbilt.edu}

\begin{abstract}
Scientific discovery in data-rich domains is currently constrained by human bandwidth: the growth in the volume and complexity of real-world data far outpaces the rate at which researchers can read, reason, and synthesize. Recent LLM-based multi-agent systems have begun to automate portions of the research cycle, but they target hypothesis generation in settings where validation cannot itself be automated, and each run is one-shot, with no mechanism for findings to accumulate or steer subsequent inquiry. This paper introduces \sysname, a multi-agent framework for autonomous knowledge discovery that is both computational and cumulative, allowing validated findings to persist and inform subsequent inquiry. Six coordinated LLM agents collaborate in an open-ended discovery loop, where accepted findings are stored in a persistent insight graph that serves as both long-term memory and an exploration-steering mechanism. We evaluate \sysname on three diverse datasets from two perspectives: Open-ended Quality against published findings, and Conditioned Quality via literature-derived queries. Across both evaluation perspectives, \sysname 
covers known findings and surfaces substantive discoveries that complement human-driven research. Our code is available at: \url{https://github.com/GeQinwen/AutoKD}
\end{abstract}


\received{20 February 2007}
\received[revised]{12 March 2009}
\received[accepted]{5 June 2009}

\maketitle

\section{Introduction.}
\label{sec:intro}

Scientific discovery (SD) follows a well-established cycle: researchers survey prior work, formulate hypotheses, design and conduct experiments, analyze results, and communicate findings that accumulate into a shared body of knowledge~\cite{kuhn1962structure}. This iterative SD process has driven centuries of progress, yet it is fundamentally constrained by human bandwidth — the time required to read, reason, explore, and synthesize grows far more slowly than the data and literature available to work with~\cite{hey2009fourth}. Recent advances in LLMs have opened the possibility of automating substantial portions of this cycle. Systems such as The AI Scientist~\cite{lu2024ai, yamada2025ai}, the AI Co-Scientist~\cite{gottweis2025towards}, AI-Researcher~\cite{tang2025ai}, and Kosmos~\cite{mitchener2025kosmos} have shown that LLM-based agents can generate research ideas, execute experiments, and even produce manuscripts with minimal human intervention. 

\begin{figure}[t]
    \centering
    \includegraphics[width=0.85\columnwidth]{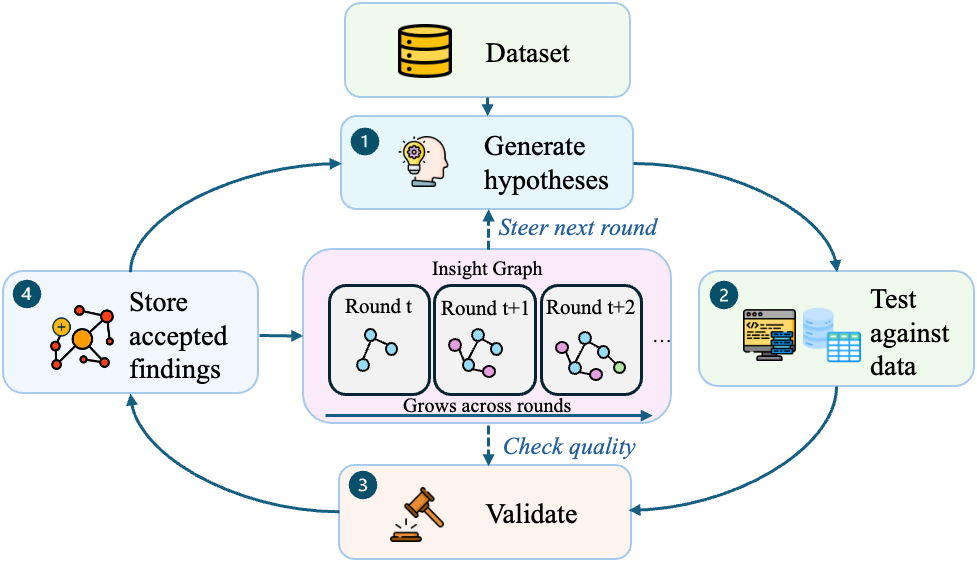}
    \vspace{-2.25ex}
    \caption{High-level overview of \sysname for autonomous knowledge discovery including insight graph storage allowing cross-round accumulation.
    }
    \label{fig:autokd_highlevel}
    \vspace{-3.5ex}
\end{figure}
 
However, these SD systems are mostly geared towards hypothesis generation and the validation step they rely on is not always amenable to automated experimentation whereas we focus on automated knowledge discovery, which is inherently computational and data-driven. Furthermore, these systems are mostly one-shot: they concentrate significant effort on producing a single paper or refining a single hypothesis, and the knowledge generated in one run does not carry forward. In contrast, real scientific progress is \emph{cumulative} — each discovery reshapes the landscape of what is known and what is worth investigating next.

What makes the full discovery process difficult to automate? Prior work in Automated Machine Learning (AutoML) has shown that complex ML workflows can be successfully automated when the objective is well defined~\cite{he2021automl}. This raises a broader question: can the process of knowledge discovery itself be automated? Unlike AutoML, knowledge discovery has traditionally relied on substantial human judgment to determine which questions are worth pursuing and which findings are meaningful, without a clear scalar objective to optimize. Recent advances in LLMs, particularly their ability to read and summarize scientific findings~\cite{lala2023paperqa}, translate natural-language hypotheses into executable statistical code~\cite{hong2025data}, and assess whether candidate findings are plausible, novel, or redundant~\cite{si2024can}, may provide the missing judgment layer needed to close this loop as shown in Figure~\ref{fig:autokd_highlevel}. Because LLMs are language-native, they can be combined with the statistical machinery already available for empirical testing, to make autonomous knowledge discovery increasingly feasible as a closed-loop process.

We therefore study the following problem: given a dataset and optionally a high-level goal, autonomously produce a cumulative and validated body of empirical findings across multiple rounds of inquiry. There are two coupled sub-problems. \emph{Per-round discovery} requires generating candidate findings, testing them against the data, and validating them without a human in the critical path. \emph{Cross-round accumulation} requires retaining and organizing prior findings so that they can guide subsequent inquiry, reduce redundancy, and encourage exploration of under-covered regions. The accumulation challenge is what distinguishes our \sysname from one-shot generation: accumulated findings must not only persist, but also be structured in a way that allows the system to reason over what is already known and determine where to explore next.

We present \textbf{\sysname}, a framework for autonomous knowledge discovery organized around discovery and accumulation. Per-round discovery is carried out through an iterative agentic loop that generates candidate findings, tests them against real data, and evaluates their validity. Cross-round accumulation is supported by a persistent \emph{insight graph} that stores accepted findings, filters candidates that duplicate existing knowledge, and steers subsequent rounds toward under-explored regions. Together, these components enable a discovery process in which knowledge is continuously accumulated and used to guide future inquiry.

Evaluating a SD system is itself a challenge, since autonomous discovery provides no ground-truth label for whether a finding is scientifically worthwhile. We therefore use published findings on the same datasets as a reference point and evaluate the resulting insight graph from two complementary perspectives: \emph{open-ended quality}, which asks whether individual discoveries are comparable to findings reported by human researchers, and \emph{conditioned quality}, which asks whether insights accumulated across rounds collectively cover findings documented in the literature.
Our contributions are: 
\begin{itemize}[leftmargin=8pt]
    \item We introduce and formalize \emph{autonomous agentic knowledge discovery}: given a dataset, the goal is to autonomously produce a cumulative and validated body of empirical findings across multiple rounds of inquiry, rather than a single paper or hypothesis as in prior AI scientist systems.

    \vspace{0.75ex}
    \item We propose \sysname, which couples an iterative discovery loop with a persistent \emph{insight graph} that accumulates validated findings across rounds, steers exploration toward under-covered regions of the hypothesis space, and filters redundant candidates so that newly accepted insights add substantive knowledge.

        \vspace{0.75ex}
    \item We propose a literature-grounded evaluation framework that assesses both individual insight quality and cumulative knowledge coverage against published findings, and use it to conduct, to the best of our knowledge, the first large-scale empirical study of autonomous knowledge discovery across three diverse datasets.

\end{itemize}

\vspace{-0.5ex}
\section{Background.}
\label{sec:background}

We review four directions of prior work that together motivate \sysname: AutoML automates pipelines with scalar objectives; classic knowledge discovery uses data mining methods to extract patterns, some of which might be interesting or useful, but typically involves humans in the loop to determine which method to use and make judgments/distinctions; AI scientist systems automate end-to-end research but are mostly one-shot and geared toward hypothesis generation in domains where validation is not amenable to computation; and agentic AI, driven by LLMs, now supplies the missing judgment layer. \sysname exploits these capabilities to make discovery cumulative and data-driven.

\vspace{-0.75ex}
\paragraph{\textbf{Automated Machine Learning}}
AutoML has successfully mechanized key stages of the ML workflow, from hyperparameter optimization and feature selection to model selection and neural architecture search~\cite{he2021automl, elsken2019neural, snoek2012practical}. AutoML succeeds because its objective is well-defined: minimize a validation loss or maximize a held-out metric. Given a clear reward signal, automated search and optimization methods can efficiently search the pipeline's configuration space without human judgment. This formulation does not directly extend to earlier conceptual stages of discovery, such as deciding what question to ask or what constitutes a meaningful finding, for which no comparable scalar objective exists.

\vspace{-0.75ex}
\paragraph{\textbf{Knowledge Discovery}}
Classical knowledge discovery in databases and data mining developed rich toolkits for surfacing structure in data, including association-rule mining, subgroup discovery, clustering, and anomaly detection~\cite{fayyad1996data, han2022data}. These methods can enumerate large numbers of statistical patterns from a dataset, but turning patterns into insights has historically required HIL --- both to choose which method fits the question and to judge whether a mined pattern is non-trivial, substantively interesting, and consistent with prior understanding~\cite{geng2006interestingness}. As a result, the full loop from raw data to validated, communicable insight has resisted AutoML-style automation. The bottleneck is not pattern extraction but the surrounding judgment: deciding which method to apply, which patterns are worth keeping, how findings relate to what is already known, and where to explore next.

\vspace{-0.75ex}
\paragraph{\textbf{AI Scientific Discovery Systems}}
\label{sec:ai-scientist}
A growing body of work operationalizes end-to-end AI scientist systems: \emph{The AI Scientist}~\cite{lu2024ai,yamada2025ai} for machine-learning research, the \emph{AI Co-Scientist}~\cite{gottweis2025towards} for biomedical hypothesis generation via tournament-style agent ranking, \emph{AI-Researcher}~\cite{tang2025ai} for full-pipeline manuscript production, and \emph{Agent Laboratory}~\cite{schmidgall2025agent} for human-in-the-loop research assistance. \emph{Kosmos}~\cite{mitchener2025kosmos} comes closest to our data-driven discovery setting, using a shared world model to coordinate parallel data-analysis and literature-search agents across long-horizon runs. Yet most of these systems concentrate effort on hypothesis generation in domains where validation cannot be automated, and they are largely one-shot: each run produces a single paper or refined hypothesis, with no mechanism for findings to accumulate or steer the next round.

\vspace{1.25ex}
\textbf{Agentic AI-Enabling Automated Knowledge Discovery.}
\label{sec:llm-kd}
The LLM agent's capabilities are exactly the judgment layer that classical knowledge discovery has lacked. LLMs can read and summarize scientific findings~\cite{lala2023paperqa}, translate natural-language hypotheses into executable statistical code~\cite{chen2021evaluating, hong2025data}, reason about whether a candidate finding is plausible, novel, or redundant relative to prior work~\cite{si2024can}, and produce interpretable explanations of quantitative results~\cite{hong2025data}. Because these capabilities are language-native, they compose naturally with the structured pipelines already available for statistical testing and data mining.

LLMs allow users to not merely generate hypotheses or summarize results in isolation, but to close the loop: an LLM-driven system can propose candidate findings, test them against data, evaluate their statistical validity and novelty, and update its own working knowledge accordingly — with no human in the critical path. This reframes automated knowledge discovery from pattern enumeration into an iterative reasoning process.

\section{Problem Definition}
\label{sec:problem}

We consider an autonomous knowledge discovery setting in which a system must iteratively discover, validate, and organize knowledge from data, potentially guided by a high-level objective.

\begin{definition}[\textbf{Autonomous Agentic Knowledge Discovery}]
Let $\mathcal{D}$ be a dataset, and optionally let $g$ be a high-level discovery objective expressed in natural language. The goal is to design an autonomous system that, given $\mathcal{D}$ (and 
$g$), produces a set of validated empirical insights $\mathcal{I}$, where each $i \in \mathcal{I}$ is a directional, testable finding supported by evidence from $\mathcal{D}$.
\end{definition}

A key characteristic of autonomous knowledge discovery is that discovery is inherently iterative and cumulative. Rather than producing a single hypothesis or report, the system operates over multiple discovery rounds, yielding a sequence of insight sets
$\mathcal{I}^{(1)} \subseteq \mathcal{I}^{(2)} \subseteq \cdots \subseteq \mathcal{I}^{(R)}$,
where insights accepted in round $r$ influence the hypotheses explored in round $r+1$. This formulation gives rise to two coupled challenges. First, \emph{per-round discovery} requires the system to generate, test, and validate candidate insights grounded in $\mathcal{D}$. Second, \emph{cross-round accumulation} requires retaining and organizing prior findings so that they can guide subsequent inquiry and reduce redundant exploration. Because only a small fraction of the hypothesis space can be tested, allocating discovery effort to one direction necessarily comes at the opportunity cost of not exploring others. Autonomous knowledge discovery therefore faces an exploration--exploitation trade-off: whether to deepen promising existing findings or explore directions that remain insufficiently sampled.

Cross-round accumulation is therefore not merely a memory problem: prior findings must inform how limited discovery effort is allocated across subsequent rounds. Addressing this accumulation and allocation challenge requires a persistent and structured representation of knowledge accumulated across rounds. Such a representation must support retrieval and comparison of prior findings, while providing the context needed to determine which directions merit further refinement and which remain under-explored.

\vspace{-1ex}
\section{\sysname Framework.}
\label{sec:method}

\begin{figure}[!t]
\centering
\includegraphics[width=0.5\textwidth]{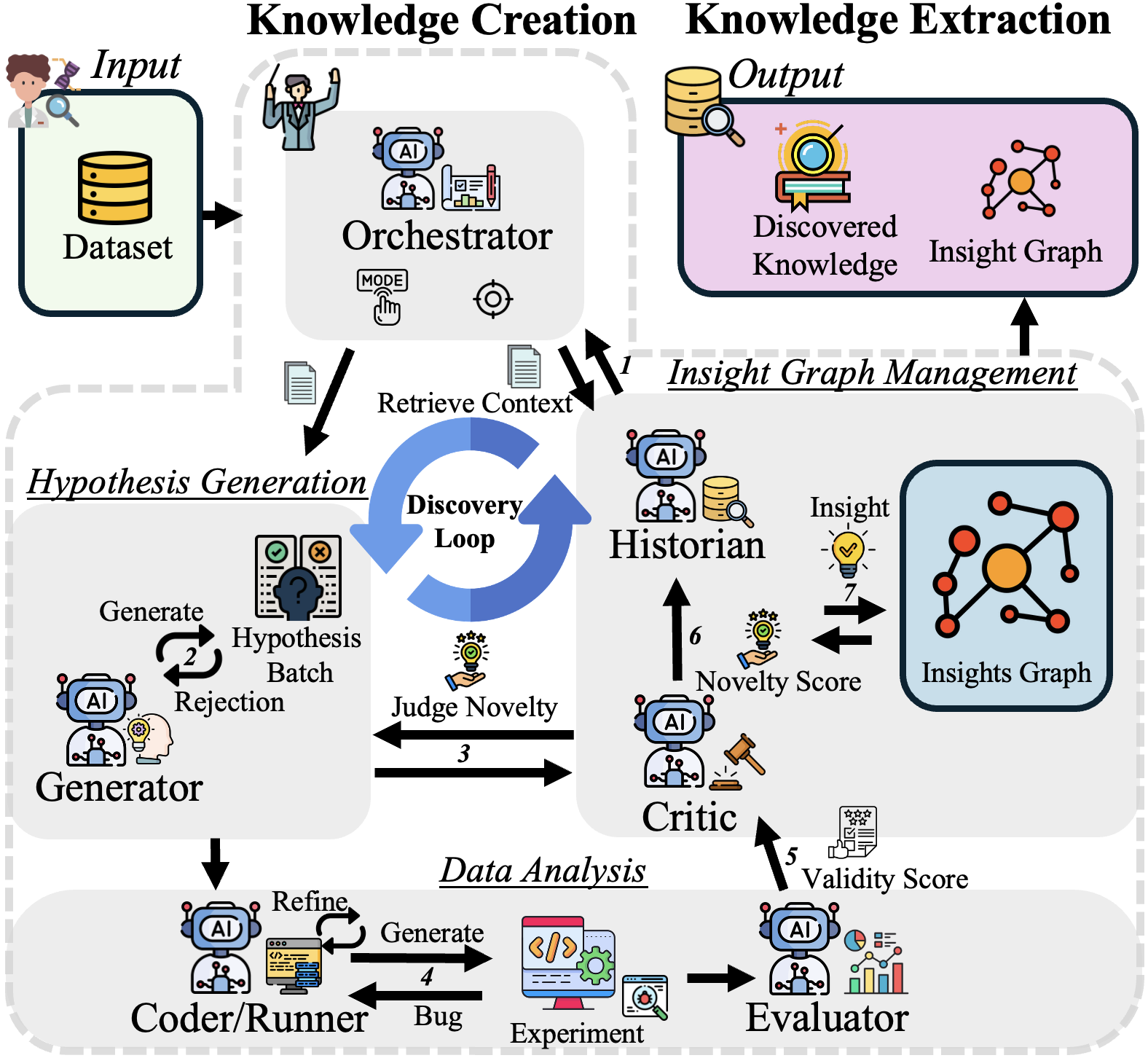}
\vspace{-3.5ex}
\caption{An overview of the \sysname framework 
for autonomous agentic knowledge discovery. \sysname first iteratively discovers and accumulates structured insights via the knowledge creation component, and then through a retrieval, synthesis, verification process enables knowledge extraction.}
\label{fig:framework}
\vspace{-3ex}
\end{figure}

\sysname takes as input a dataset $\mathcal{D}$ and optionally a discovery goal $g$ specified in natural language, and produces a persistent insight graph of validated empirical findings. The framework consists of two complementary components: \emph{knowledge creation} (Section~\ref{sec:knowledge-creation}), which iteratively discovers and accumulates insights, and \emph{knowledge extraction} (Section~\ref{sec:extraction}), which retrieves and synthesizes accumulated knowledge to answer research queries.
Figure~\ref{fig:framework} illustrates the overall architecture.

We represent the system state as an insight graph $\mathcal{G}=(\mathcal{I},\mathcal{E})$, where $\mathcal{I}$ is the set of validated insight nodes and each edge $(i_a,i_b,t)\in\mathcal{E}$ connects two insights with relation type $t\in\mathcal{T}$, where $\mathcal{T}=\{\textsc{Deepens},\textsc{Extends}, \textsc{Narrows},\textsc{Contradicts}\}$ (see Appendix~\ref{app:rationale-edges}). At discovery round $r$, the current graph conditions what the system explores: \begin{align} \mathcal{H}^{(r)} &= \operatorname{Propose} \bigl(\mathcal{D},g,\mathcal{G}^{(r)}\bigr), \label{eq:propose}\\ \mathcal{A}^{(r)} &= \left\{ h\in\mathcal{H}^{(r)} : \operatorname{Valid}(h,\mathcal{D}) \land \operatorname{Novel}(h,\mathcal{G}^{(r)}) \right\}, \label{eq:accept}\\ \mathcal{G}^{(r+1)} &= \operatorname{Update} \bigl(\mathcal{G}^{(r)},\mathcal{A}^{(r)}\bigr). \label{eq:update} \end{align} Here $\mathcal{H}^{(r)}$ denotes the candidate hypotheses proposed in round $r$, and $\mathcal{A}^{(r)}$ the accepted findings. Thus, the insight graph is not only a persistent record of discovered knowledge, but the evolving state that guides subsequent inquiry. Agent details and orchestration modes appear in Appendix~\ref{sec:agents} and~\ref{app:rationale-modes}.

Six specialized agents coordinate the discovery insight loop (see Figure~\ref{fig:framework}), each responsible for a distinct stage of the discovery cycle.
Each agent $A_k$ can be characterized by its input--output signature: $A_k : \mathcal{X}_k \rightarrow \mathcal{Y}_k$, where $\mathcal{X}_k$ and $\mathcal{Y}_k$ denote the agent's input and output spaces, respectively. As shown in Figure~\ref{fig:framework}, \sysname takes a dataset and an optional discovery goal as input and produces a growing insight graph as output: (1)~the \emph{Orchestrator} plans each round by consulting the Historian for context from the insight graph; (2)~the \emph{Generator} produces a batch of candidate hypotheses; (3)~the \emph{Critic} filters redundancy for novelty; (4)~the \emph{Coder/Runner} translates hypotheses into executable code and runs them against data; (5)~the \emph{Evaluator} scores statistical validity; (6)~the \emph{Critic} assesses novelty; and (7)~the \emph{Historian} stores accepted insights with typed relations in the insight graph, which steers subsequent rounds. The paper introduces the overall discovery loop and the insight graph structure, where details of each agent are described in Appendix~\ref{sec:agents}.

{
\SetAlgoSkip{tightalgoskip}
\begin{algorithm}[t]
\caption{\sysname Discovery Loop}
\label{alg:discovery}
\small 
\KwIn{dataset $\mathcal{D}$, goal $g$, max rounds $R$, stall window $\tau$}
\KwOut{insight graph $\mathcal{G}$}
$\mathcal{G} \gets \emptyset$,\ $\mathcal{J} \gets \emptyset$,\ $\mathcal{F} \gets \emptyset$,\ $c \gets 0$\;
\For{$r = 1$ \KwTo $R$}{
  $(m_r, \mathbf{a}_r) \gets \textsc{Orchestrator.Plan}(\mathcal{G}, g)$\;
  $\mathcal{C}_r \gets \textsc{Historian.GetContext}(\mathbf{a}_r, \mathcal{G})$\;
  $\mathcal{H}_r \gets \textsc{Generator.Propose}(m_r, \mathcal{C}_r, g, \mathcal{D}; \mathcal{F}, \mathcal{J})$\;
  $\mathcal{H}_r^{+} \gets \textsc{Critic.Filter}(\mathcal{H}_r, \mathcal{G})$\;
  add rejected hypotheses from $\mathcal{H}_r \setminus \mathcal{H}_r^{+}$ to $\mathcal{J}$\;
  \ForEach{$h \in \mathcal{H}_r^{+}$}{
    $\rho_h \gets \textsc{CoderRunner.Execute}(h, \mathcal{D})$ \tcp*{retry $\leq k$}
    \If{$\rho_h \neq \bot$}{
      $s_v(h) \gets \textsc{Evaluator.Score}(\rho_h, \mathcal{D})$\;
      $(s_n(h), d) \gets \textsc{Critic.Judge}(h, \rho_h, s_v(h), \mathcal{G})$\;
      \uIf{$d = \textsc{Accept}$}{
        insert node $(h, \rho_h, s_v, s_n)$ into $\mathcal{G}$\;
        $\textsc{Historian.Classify}(h, \mathcal{G})$\;
        $n_{\text{new}} \mathrel{+}= 1$\;
      }
      \uElseIf{$d = \textsc{Refine}$}{
        add $(h, \rho_h, \text{hint})$ to $\mathcal{F}$\;
      }
      \Else{
        add $(h, \rho_h, \text{reason})$ to $\mathcal{J}$\;
      }
    }
  }
  $c \gets (n_{\text{new}} = 0)\;?\; c+1 : 0$ \tcp*{stop if $\tau$ empty rounds}
  \lIf{$c \geq \tau$}{\textbf{break}}
}

\end{algorithm}
}

\subsection{Discovery Loop for Knowledge Creation} 
\label{sec:knowledge-creation}

The knowledge creation process operates as a multi-round discovery loop. In each round, \sysname plans what to investigate, generates
candidate hypotheses, tests hypotheses against data, evaluates hypotheses' statistical validity, judges their novelty relative to existing knowledge, and stores accepted findings in the insight graph.

\subsubsection{\textbf{Discovery Loop}}
Figure~\ref{fig:framework} annotates the data flow with step numbers to show how the agents compose into a single discovery round, while also providing details in Algorithm~\ref{alg:discovery}.

\paragraph{Cold start.}
When the insight graph is empty, a deterministic seeding pass
populates $\mathcal{G}^{(0)}$ with two kinds of orientation nodes before round~1 begins: (i) descriptive seeds: carrying basic statistics for each thematic module in $\mathcal{D}$, and (ii)cross-module briefs: summarizing the strongest inter-module correlations. These seeds and briefs give the Orchestrator an initial map of the dataset's landscape.

\smallskip\noindent\textit{Stage 1: Hypothesis Generation.}
\vspace{-1ex}
\paragraph{Planning and generation.}
The Orchestrator selects an exploration mode and a set of anchor insights from $\mathcal{G}$.  The Historian assembles round context insights $\mathcal{C}_r$ by retrieving the anchors and their graph neighborhood.  The Generator, conditioned on this context, the discovery goal, and any previously rejected hypotheses, runs a two-phase flow: it first picks a thematic direction from a concept menu, then refines it into a batch of $B$ testable hypotheses using variable-level metadata retrieved from a structured catalog.

\paragraph{Pre-execution filter.}
The Critic inspects each candidate against $\mathcal{G}$, removing hypotheses that are structurally duplicate (overlapping core variables with an existing node), semantically duplicate (high embedding similarity), or tautological.  Rejected candidates are logged to $\mathcal{J}$ with filter reasons.

\vspace{1ex}
\smallskip\noindent\textit{Stage 2: Data Analysis.}
\vspace{-1ex}
\paragraph{Execution and evaluation.}
Each surviving hypothesis is compiled into executable code by the Coder/Runner. The Evaluator then maps the execution output $\rho_h$ to a composite validity score $s_v(h)$, applying penalties for trivially small effect sizes and split-sample instability, and translates numerical results into natural-language form for the Critic to assess the quality of the insights.

\vspace{1ex}
\smallskip\noindent\textit{Stage 3: Insight Graph Management.}
\vspace{-1ex}
\paragraph{Judgment and storage.}
The Critic's post-execution judge computes a novelty score $s_n(h)$ blending structural distance, semantic distance, and an LLM-assessed epistemic surprise component.  Based on the combined quality signal, it assigns a disposition: \textsc{Accept}, \textsc{Refine}, or \textsc{Reject}.  Accepted insights are handed to the Historian, which inserts the new node into $\mathcal{G}$ and classifies typed edges (e.g. \textsc{Extends}) to its anchors.

The updated graph steers the next round: the Orchestrator's mode weights shift as coverage grows, and the Critic's duplicate thresholds tighten as semantically adjacent nodes accumulate. The loop continues until $\tau$ consecutive rounds produce no new accepted insights.

\subsubsection{\textbf{Insight Graph.}}
\label{sec:insight_graph}

The insight graph $\mathcal{G} = (\mathcal{I}, \mathcal{E})$ is system 
persistent memory. Each node $i \in \mathcal{I}$ stores a 
finding as a tuple:
\begin{align*}
i = (h,\; \rho_h,\; s_v(h),\; s_n(h)),
\end{align*}
where $h$ is the hypothesis specification,  $\rho_h$ is the execution result, $s_v(h)$ and $s_n(h)$ are the validity and novelty scores.
 Each edge $(i_a, i_b, t) \in \mathcal{E}$ carries a relation type $t \in \mathcal{T}$ assigned by the Historian during insertion. The graph grows monotonically across rounds: $\mathcal{G}^{(r+1)} \supseteq \mathcal{G}^{(r)}$.

\paragraph{Relation to knowledge graphs.} Although $\mathcal{G}$ borrows the node-and-edge structure of a knowledge graph, its semantics differ. In a conventional knowledge graph, nodes denote entities and edges denoting factual relations between them. In the \emph{insight graph}, each node is itself an empirically validated proposition, i.e. a hypothesis paired with its execution result and validity/novelty scores, as well as edges capturing \emph{epistemic} relations between findings. This shift from entity-level to finding-level representation is what lets $\mathcal{G}$ serve as \emph{cumulative discovery memory}: retrieval, novelty judgment, and gap-driven exploration all operate over validated claims rather than atomic facts.

\subsection{Knowledge Extraction}
\label{sec:extraction}

While knowledge creation populates the insight graph $\mathcal{G}$, knowledge extraction enables querying and evaluation of $\mathcal{G}$'s contents. Given a natural-language 
question $q$, the extraction pipeline retrieves relevant insights from $\mathcal{G}$ and synthesizes a grounded answer through a three-stage process: retrieval, synthesis, and verification.

\paragraph{Retrieval.}
The system embeds the query to $\mathbf{e}_q$ and selects seed nodes by cosine similarity against all insight embeddings. One-hop neighbor expansion from the seeds along graph edges gathers structurally related candidates. Each candidate is scored by a weighted combination of semantic similarity and stored validity, and the top-$K$ form the retrieved set $\mathcal{I}_q$.
\begin{align*}
\textsc{Retrieve} : (q,\; \mathcal{G}) \;\longrightarrow\; \mathcal{I}_q
\end{align*}

\paragraph{Synthesis.}
A synthesizer LLM receives the query and retrieved insights with their node identifiers, and generates a natural-language answer $a_q$ that grounds every claim in a cited node. Citations are extracted to produce $\mathcal{C}_q = \{(c_j, I_{c_j})\}$, mapping each citation to its supporting insight in $\mathcal{I}_q$.
\vspace{-0.5ex}
\begin{align*}
\textsc{Synthesize} : (q,\; \mathcal{I}_q) \;\longrightarrow\; (a_q,\; \mathcal{C}_q)
\end{align*}

\vspace{-2ex}
\paragraph{Verification.}
A separate judge LLM decomposes $a_q$ into individual claims and checks each against its cited evidence via binary entailment. The faithfulness score $\mathrm{Faith}(q)$ is the fraction of supported claims; $\mathrm{Halluc}(q) = 1 - \mathrm{Faith}(q)$. Unsupported assertions are flagged to ensure the answer reflects the graph contents rather than hallucinated knowledge.
\vspace{-0.5ex}
\begin{align*}
\textsc{Verify} : (a_q,\; \mathcal{C}_q,\; \mathcal{I}_q) \;\longrightarrow\; \big(\mathrm{Faith}(q),\; \mathrm{Halluc}(q)\big)
\end{align*}
\section{Proposed Literature-Grounded Evaluation for Autonomous Knowledge Discovery}
\label{sec:eval-framework}

\begin{figure}[!t]
\centering
\includegraphics[width=0.46\textwidth]{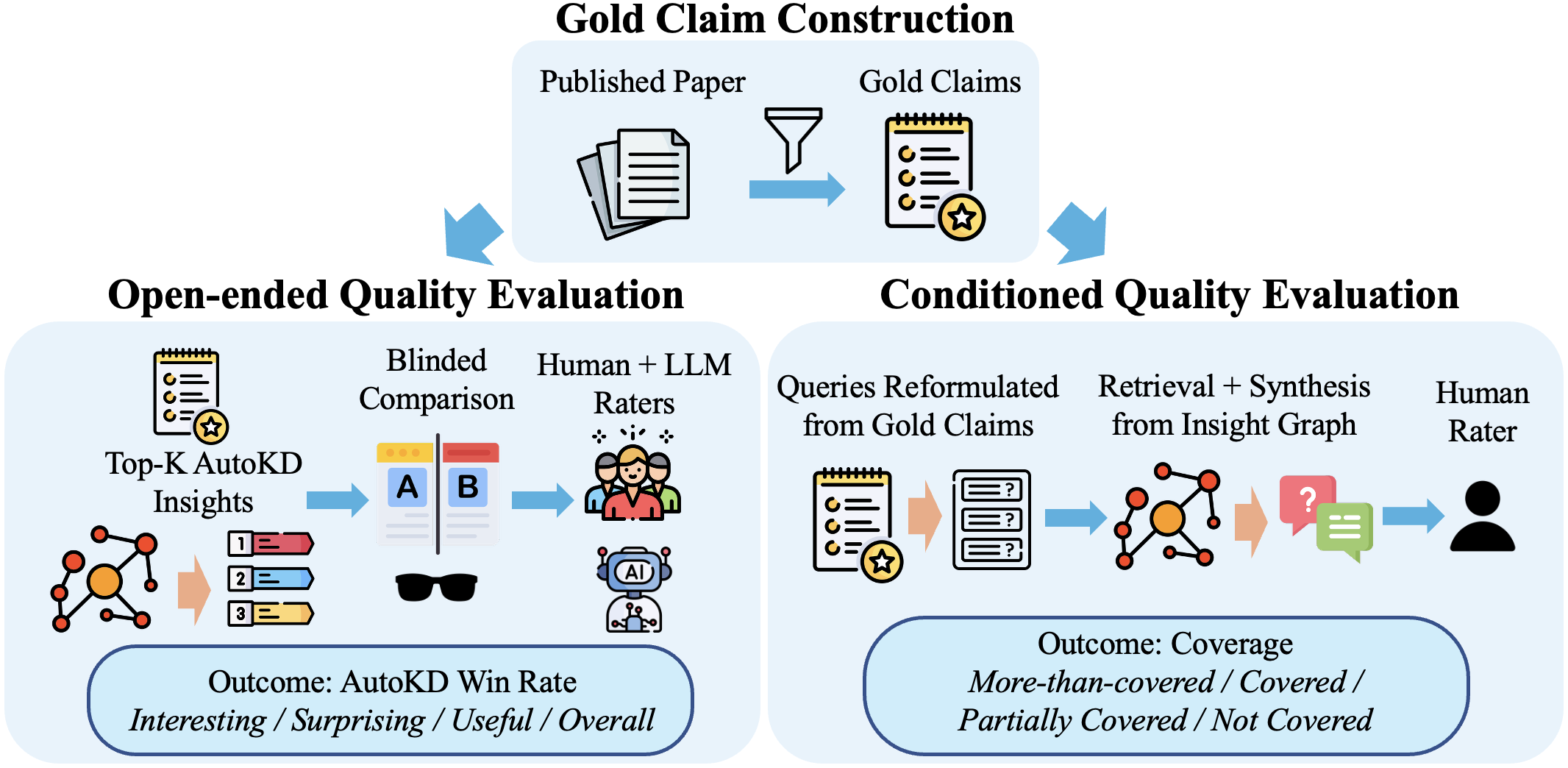}
\vspace{-2ex}
\caption{Proposed evaluation framework for autonomous knowledge discovery. Gold claims distilled from published papers support two complementary protocols: \emph{open-ended quality} compares \sysname insights with gold claims, while \emph{conditioned quality} evaluates literature-derived queries through retrieval and synthesis over the insight graph.}
\label{fig:eval}
\vspace{-2.5ex}
\end{figure}

Open-ended discovery lacks predefined ground truth, making evaluation itself a challenge. We therefore propose a literature-grounded framework that uses published findings on the same datasets to assess two complementary capabilities:

\begin{itemize}
\item 
\textbf{RQ1. Single-insight Quality} To what extent can an 
autonomous knowledge discovery system
discover individual insights from $\mathcal{D}$ that are substantively comparable to findings reported in published studies on the same data?
\item 
\textbf{RQ2. Multi-insight Coverage} 
To what extent do insights accumulated across rounds collectively cover known findings from the literature, such that literature-grounded research questions can be answered from the insight graph?
\end{itemize}

To answer RQ1, the \emph{open-ended quality evaluation}
(Section~\ref{sec:eval-open}) pairs the system's top-ranked discoveries with
matched gold claims from published work for blinded head-to-head comparison.
To answer RQ2, the \emph{conditioned quality evaluation}
(Section~\ref{sec:eval-cond}) instantiates a GraphRAG interaction~\cite{perozzi2024let}
over the insight graph: research questions derived from published findings are
answered by retrieving relevant nodes and synthesizing grounded responses from
the graph. Both evaluations use a common set of gold claims curated from papers
analyzing the same data source (Section~\ref{sec:gold-claim}).
Figure~\ref{fig:eval} summarizes the framework.

\subsection{Open-ended Quality Evaluation}
\label{sec:eval-open}
\textit{Top-\textit{K} Set Selection.}
For each dataset, \sysname ranks all validated insights based on the novelty and validity scores assigned during knowledge creation. A diversity filter selects the top-10 insights based on re-ranking, skipping any candidate whose cosine similarity to an already-selected insight exceeds a threshold, ensuring the relative diversity of the final set. Separately, 10 representative gold claims are drawn from the curated set. The two lists are aligned using Hungarian matching on sentence-transformer cosine similarity to produce the most favorable pairing. The full 30 pairs shown to evaluators are reported in Supplementary S3.

\textit{Raters.}
The open-ended evaluation uses eight raters: four human raters with graduate-level research training and four LLM judges (\texttt{claude-opus-4-7}, \texttt{Gemini-3}, \texttt{DeepSeek-V3}, \texttt{gpt-5.5}). The blinded presentation described in Appendix~\ref{app:human-comparison} prevents raters from knowing which finding in any given pair was system-generated. All raters received only the written instrument shown on the form.

\textit{Rating instrument.}
For each pair, human raters answer four forced-choice questions with options \{Finding A, Finding B, Tie\}: \emph{More interesting} (which finding the rater would be more curious to read about or explore further), \emph{More surprising} (which is more unexpected given prior knowledge or intuition of the domain), \emph{Most useful} (which is more likely to guide future research or real-world decision-making), and \emph{Overall better} (a holistic judgment of scientific strength).

For each LLM judge, pairs are submitted one at a time rather than as the full batch of 30, so that judgments on one pair cannot be anchored by judgments on adjacent pairs. The prompt presents the dataset context, 
two findings, and the four questions in the same wording shown to human raters, and constrains the output to a structured response covering all four dimensions. All 
LLM judges receive the same 
A/B assignment used in the human evaluation. 

For each dimension we report the percentage of non-tie responses in which \sysname is preferred over the matched gold claim. Ties are excluded from the denominator.

\vspace{-1ex}
\subsection{Conditioned Quality Evaluation}
\label{sec:eval-cond}

For each dataset, five diverse gold claims are sampled from $\mathcal{G}^\star$ and reformulated as natural-language research questions. The extraction pipeline
(Section~\ref{sec:extraction}) retrieves relevant insights from $G$ and synthesizes a grounded answer per query.

A human rater with graduate-level research training scores each query outcome on a four-level ordinal scale: \emph{More-than-covered (MC)}: the graph contains the claim plus additional consistent evidence or detail that extends it; \emph{Covered (C)}: a directional match on the same variables; \emph{Partially covered (P)}: a related finding sharing some but not all of the claim's variables or conditions, or matching direction on a proxy; \emph{Not covered (NC)}: no matching or related finding is retrievable.

\vspace{-1ex}
\subsection{Gold Claim Construction}
\label{sec:gold-claim}

For each dataset, we curate a reference set $\mathcal{G}^\star = \{g_1, \ldots, g_M\}$ of gold claims from published research that analyzes the same data source. The curation follows the following process: First, we retrieve papers citing the dataset's primary publication on Google Scholar and filter by the dataset name as a keyword to retain only papers that use the dataset as a primary data source rather than as background reference. We paginate through results until a sufficient pool of qualifying papers is collected. Second, we screen each candidate paper to verify that it conducts empirical analysis on the dataset and reports findings that are in principle testable by our system. Third, we extract directional empirical claims from the passing papers, splitting compound findings into individual sub-claims and lightly rephrasing for consistency without altering meaning. Finally, evaluation queries are constructed from these claims so that each query $q$ is annotated with a subset $\mathcal{G}^\star_q \subseteq \mathcal{G}^\star$ of claims it should address. See Appendix~\ref{app:human-comparison} for the blinded comparison protocol.

\vspace{-0.75ex}
\section{Experiments}
\label{sec:experiments}

We show empirical evaluations of \sysname under our framework introduced in Section~\ref{sec:eval-framework} across three datasets spanning distinct discovery domains. We first describe the experimental setup, then report results for open-ended quality and conditioned quality, followed by further analysis of the accumulated insight graph.

\vspace{-0.75ex}
\subsection{Experimental Setup}

\subsubsection{Agent Setup}
All in-loop agents share a single open-weight backbone, \texttt{Qwen~2.5-14B} (Q4\_K\_M). Embeddings for retrieval and diversity filtering use \texttt{all-MiniLM-L6-v2}~\cite{reimers2019sentence}. Generation uses temperature $0.75$; deterministic agents (Critic judge, Evaluator, LLM judges) use $T{=}0$. Discovery runs on a single workstation (RTX~4090, 24\,GB) with an early-stop window of $\tau{=}30$ empty rounds (no fixed round budget; the loop terminates when discovery saturates). Hyperparameters are in Table~7; prompts are in Supplementary S2.

Although the Coder/Runner design in Section~\ref{sec:agents} can freely translate any hypothesis into executable code, our current \sysname implementation restricts execution to four templates in a statistical domain-specific language (\texttt{assoc}, \texttt{diff}, \texttt{interact}, \texttt{heterogeneity}) that compile deterministically (Supplementary S1). This eliminates the estimator hallucinations and run-to-run drift of free-form code generation, and already covers the bulk of testable findings on our datasets --- even so, the hypothesis space is combinatorially vast (Appendix~\ref{app:hypothesis-space}), making selection rather than enumeration the binding challenge. The fully general design awaits LLM coding agents capable of producing audit-grade statistical code.

\subsubsection{Datasets}
To test generalization across substantively different discovery settings, we select three datasets spanning distinct domains: e-commerce (\emph{Amazon Reviews}~\cite{ni2019justifying}), cross-national social science (\emph{WVS Wave~7}~\cite{haerpfer2022world}), and bibliometrics (\emph{SciSciNet~v2}~\cite{lin2023sciscinet}). 

\vspace{-1ex}
\paragraph{Amazon Reviews.}
The Amazon Review Data~\cite{ni2019justifying} is a large-scale e-commerce dataset. We use the Books 5-core subset joined with product metadata, yielding 34 analysis-ready variables covering ratings, helpfulness votes, product attributes (price, sales rank, category), reviewer behavior, and simple review NLP text features. A 500K-row random sample is drawn at discovery time for computational feasibility.

\vspace{-1ex}
\paragraph{World Values Survey.}
WVS Wave~7~\cite{haerpfer2022world} is a cross-national public opinion survey, including more than 80,000 respondents across 64 countries and territories. The 206 variables include opinion and demographic items spanning social values, political attitudes, trust, well-being, gender norms, religion, and security, along with survey design weights, country-level macro indicators (GDP, urbanization, death rate), and derived composite indices.

\vspace{-1ex}
\paragraph{SciSciNet.}
SciSciNet~v2~\cite{lin2023sciscinet} is a large-scale science of science dataset comprising papers, authors, citations, and external linkages. We use both the computer-science subset and all-fields subset, aggregating the normalized source tables into a paper-level flat table with 88 analysis-ready variables covering citation impact, novelty and disruption, sleeping beauty indicators, team composition, author career metrics, real-world impact (patents, news, clinical trials), field interdisciplinarity, reference age profile, open access status, and text features. A 500K-row random sample is drawn at discovery time for computational feasibility.

\begin{table}[t]
\centering
\caption{\sysname win rate (\%, ties excluded) by dataset, dimension,
and rater subset. Human and LLM results each aggregate four raters;
All aggregates all eight raters.}
\label{tab:win-rate-breakdown}
\vspace{-2ex}
\small
\setlength{\tabcolsep}{5pt}

\begin{tabular}{llrrrr}
\toprule
\textbf{Dataset} & \textbf{Rater}
& \textbf{Int.} & \textbf{Surp.}
& \textbf{Useful} & \textbf{Overall} \\
\midrule

\multirow{3}{*}{WVS}
& Human & 83.8 & 86.1 & 81.1 & 86.1 \\
& LLM   & 97.4 & 100.0 & 66.7 & 92.5 \\
& \textbf{All} & 90.7 & 93.2 & 73.7 & 89.5 \\

\midrule
\multirow{3}{*}{Amazon}
& Human & 53.1 & 56.2 & 63.6 & 54.5 \\
& LLM   & 50.0 & 56.4 & 17.5 & 47.5 \\
& \textbf{All} & 51.4 & 56.3 & 38.4 & 50.7 \\

\midrule
\multirow{3}{*}{SciSciNet}
& Human & 54.5 & 62.5 & 52.8 & 51.4 \\
& LLM   & 65.0 & 65.0 & 37.5 & 61.5 \\
& \textbf{All} & 60.3 & 63.9 & 44.7 & 56.8 \\

\midrule
\multirow{3}{*}{\textbf{All}}
& Human & 64.7 & 69.0 & 66.0 & 64.4 \\
& LLM   & 70.3 & 73.3 & 40.3 & 67.2 \\
\rowcolor{gray!10}
& \textbf{All} & 67.7 & 71.3 & 52.4 & 65.9 \\

\bottomrule
\end{tabular}
\vspace{-1.5ex}
\end{table}

\begin{table}[t]
\centering
\small 
\setlength{\tabcolsep}{4pt}
\renewcommand{\arraystretch}{0.92}

\caption{Inter-rater agreement (Fleiss' $\kappa$).}
\label{tab:fleiss}
\vspace{-2ex}
\begin{tabular}{lcccc}
\toprule
& \textbf{Interesting} & \textbf{Surprising} & \textbf{Useful} & \textbf{Overall} \\
\midrule
\multicolumn{5}{l}{\emph{Humans only (4 raters)}} \\
WVS          &    .012 &    .064 & $-.038$ & $-.070$ \\
Amazon       &    .139 &    .186 &    .255 &    .202 \\
SciSciNet    & $-.065$ & $-.075$ & $-.056$ & $-.053$ \\
\textbf{All} &    .098 &    .111 &    .111 &    .112 \\
\midrule
\multicolumn{5}{l}{\emph{LLM judges only (4 raters)}} \\
WVS          &    .056 & $-.081$ &    .187 &    .640 \\
Amazon       &    .733 &    .613 & $-.097$ &    .632 \\
SciSciNet    &    .853 &    .487 &    .324 &    .733 \\
\textbf{All} &    .745 &    .548 &    .308 &    .728 \\
\bottomrule
\end{tabular}
\vspace{-2ex}
\end{table}

\begin{table*}[t]
\centering
\setlength{\tabcolsep}{2.75pt}
\footnotesize 
\caption{Representative cases where all four human evaluators preferred 
the \sysname-generated finding over its gold-benchmark counterpart 
along the indicated dimension.}
\label{tab:case-examples}
\vspace{-0.75em}
\begin{tabular}{@{}p{1.5cm} p{7.2cm} p{6.6cm} p{1.5cm}@{}}
\toprule
Case & \sysname Finding (preferred) & Gold Counterpart & \hspace{-2.5ex}LLM Judges \\
\midrule
Amazon \newline \emph{\scriptsize (useful)} 
  & Longer reviews tend to appear on less popular books. 
  & Longer reviews receive more helpful votes. 
  & Disagree \\[2pt]
WVS \newline \emph{\scriptsize(interesting)} 
  & Higher social trust is negatively associated with discomfort with women earning more than their husbands. 
  & Being in a relationship is positively associated with higher life satisfaction. 
  & Agree \\[2pt]
SciSciNet \newline \emph{\scriptsize(surprising)} 
  & Heavily cited papers span fewer broad academic fields. 
  & NIH- and NSF-funded papers are more likely to receive patent citations than non-funded papers. 
  & Agree \\
\bottomrule
\end{tabular}
\vspace{-3pt}
\end{table*}

\begin{figure*}[t]
  \centering
  \includegraphics[width=\textwidth]{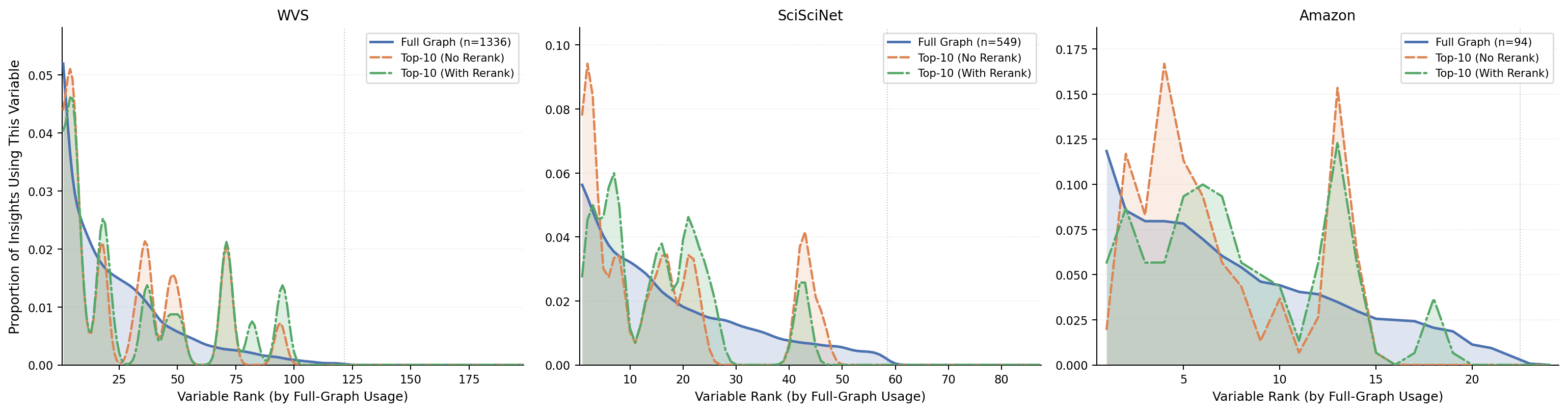}
  \vspace{-4ex}
  \caption{Variable-usage distribution for the full insight graph
    (blue), the score-only top-10 (orange), and the top-10 after
    diversity reranking (green). Variables are ordered by full-graph
    frequency.}
  \label{fig:panel-var-freq}
\vspace{-5pt}
\end{figure*}

\begin{table}
\small
\caption{Gold-claim coverage by \sysname (\%). MC/C/P/NC =
more-than covered/covered/partially covered/not covered.}
\label{tab:coverage}
\vspace{-0.5em}
\setlength{\tabcolsep}{7pt}
\begin{tabular}{lrrrr}
\toprule
& \textbf{MC} & \textbf{C} & \textbf{P} & \textbf{NC} \\
\midrule
WVS       & 0.0  & 40.0 & 60.0 & 0.0 \\
Amazon    & 40.0 & 0.0  & 60.0 & 0.0 \\
SciSciNet & 20.0 & 0.0  & 80.0 & 0.0 \\
\midrule
Avg.      & 20.0 & 13.3 & 66.7 & 0.0 \\
\bottomrule
\end{tabular}
\vspace{-1ex}
\end{table}

\subsection{Results on Open-ended Quality}

Table~\ref{tab:win-rate-breakdown} reveals three patterns:

\begin{itemize}
\vspace{-0.5ex}
\item \textbf{At- or above-parity on every dimension.}
Aggregated across raters and datasets, \sysname is preferred
over matched gold claims on all four dimensions, with the largest
margin on \emph{Surprising} where the novelty-aware scoring targets;
Table~\ref{tab:case-examples} shows representative cases where all
four human evaluators preferred the \sysname finding.

\item \textbf{LLM judges agree more but diverge on \emph{Useful}.}
The human--LLM gap is dimension-specific: negligible on
\emph{Interesting} and \emph{Surprising}, sharp on \emph{Useful}). LLM judges agree with one another more than humans
do (Table~\ref{tab:fleiss}). One reading is that LLMs weight real-world
actionability and favor canonical gold claims, while human raters weight
research guidance and favor \sysname's less-obvious associations.

\item \textbf{Wins grow with dataset complexity.}
Overall win rates follow WVS $>$ SciSciNet $>$ Amazon, tracking both variable-space size and the distance from raters' prior intuitions: Amazon's gold claims (e.g., longer reviews attract more helpful votes) are well-known folk results, i.e., a stronger baseline.

\end{itemize}

\subsection{Results on Conditioned Quality}

Due to space limit, we report the knowledge extraction result case in Appendix~\ref{app:extraction-examples}, and report coverage in Table~\ref{tab:coverage}.

\vspace{-0.5ex}
\begin{itemize}
\item \textbf{Broad thematic coverage.}
No evaluated query is rated as not covered: 33.3\% are covered or more-than-covered, while 66.7\% are partially covered. Partial coverage often reflects that the graph captures the main relationship or theme of a published finding without matching every variable or condition, indicating broad thematic overlap with the literature-grounded findings.

\item \textbf{Discovery can extend beyond reproduction.}
For 20.0\% of queries, the graph more than covers the corresponding
gold claim, providing additional details or interaction evidence
beyond the published finding. This suggests that the iterative loop
can elaborate known findings rather than only reproduce them.
\end{itemize}

\subsection{Further Investigation on Insight Graph \\and Ablation}
Table~\ref{tab:discovery} reports the accumulated insight graphs. We further
test whether diversity reranking yields a representative rather than redundant
top-10 by comparing variable-usage distributions for the full graph, score-only
top-10, and reranked top-10 (shown in Figure~\ref{fig:panel-var-freq}). Score-only
selection concentrates on a few frequent variables, whereas reranking reduces
these peaks and extends coverage to less-used variables.

\begin{table}[t]
\centering
\setlength{\abovecaptionskip}{3pt}
\setlength{\belowcaptionskip}{3pt}
\renewcommand{\arraystretch}{0.95}
\caption{Discovery output per dataset. Insights are validated findings stored in the graph; relations are typed edges connecting them.}
\label{tab:discovery}
\small   
\vspace{1.25ex}
\begin{tabular}{lccc}
\toprule
 & \textbf{Amazon} & \textbf{WVS} & \textbf{SciSciNet (CS / All)} \\
\midrule
Insights  & 119 & 1{,}367 & 565 / 1{,}027 \\
Relations & 268 & 3{,}640 & 1{,}496 / 2{,}794 \\
\bottomrule
\end{tabular}
\vspace{-3pt}
\end{table}

\paragraph{Ablation.}
We isolate three mechanisms on WVS under a matched 1,000-round budget:
\emph{w/o persistence} removes cross-round memory; \emph{w/o graph structure}
retains persistent insights but replaces graph retrieval with flat similarity
retrieval under the same context budget; and \emph{global-only} removes
graph-selected anchors and targeted follow-up. All other pipeline components
remain unchanged.

We measure accumulation using \emph{estimands per insight}, the number of
distinct tested estimands per accepted insight, and \emph{exact rediscovery},
which calculates the fraction that exactly repeat an earlier specification. \emph{Deep-test
share} is the fraction of accepted findings using interaction or heterogeneity
tests, and $\Delta$ Depth is its relative change from \sysname.

\begin{table}[t]
\centering
\small
\caption{\textbf{Ablation of cumulative discovery mechanisms on WVS at a matched 1,000-round budget.}
$\Delta$ Depth is the relative change in deep-test share from full \sysname.}
\label{tab:ablation}
\vspace{-0.5em}
\setlength{\tabcolsep}{3pt}
\renewcommand{\arraystretch}{1.0}
\begin{tabular}{lrrrrr}
\toprule
Variant
& $n$
& \shortstack{Estimands\\ / insight $\uparrow$}
& \shortstack{Exact\\ rediscovery $\downarrow$}
& \shortstack{Deep-test\\ share $\uparrow$}
& $\Delta$ Depth \\
\midrule
\textbf{\sysname}
& 819 & \textbf{0.949} & \textbf{0.0\%} & \textbf{28.9\%} & -- \\
w/o persistence
& 1,021 & 0.672 & 16.7\% & 12.5\% & $-56.7\%$ \\
w/o graph structure
& 828 & 0.944 & 0.0\% & 26.0\% & $-10.3\%$ \\
global-only
& 785 & 0.941 & 0.0\% & 20.1\% & $-30.4\%$ \\
\bottomrule
\end{tabular}
\vspace{-12pt}
\end{table}

Removing persistence causes substantial rediscovery (16.7\%) and the largest
drop in depth ($-56.7\%$). Flat persistent memory largely preserves distinctness
but reduces depth, while global-only exploration reduces it further
($-30.4\%$), showing complementary benefits from persistence, graph structure,
and graph-guided steering.
\section{Conclusion.}
\label{sec:conclusion}

We formalized \emph{automated agentic knowledge discovery}, the task of autonomously producing a cumulative, validated body of empirical findings from a dataset, and introduced \sysname, a multi-agent framework that couples an iterative discovery loop with a persistent \emph{insight graph} that accumulates findings, steers exploration toward under-covered regions, and filters duplicates. To our knowledge, this is the first empirical study of agentic knowledge discovery at scale: across three diverse datasets, \sysname's insight graph is preferred over matched gold claims on all four open-ended dimensions and returns relevant evidence for every literature-grounded query.

\appendix

\section{\sysname's Agent Descriptions}
\label{sec:agents}

\paragraph{Orchestrator.}
The Orchestrator initiates each round by selecting an exploration mode and identifying a focus area. It consults the Historian (discussed in detail below)
to retrieve anchor insights from the graph and assembles a round context that conditions all downstream generation. This steering mechanism balances deepening existing findings with exploring new territory:
\begin{align*}
A_{\text{orch}} : (\mathcal{G}^{(r)},\; g) \;\longrightarrow\; (m_r,\; \mathbf{a}_r),
\end{align*}
where $m_r \in \{\texttt{local},\, \texttt{global},\, \texttt{refine},\, \texttt{conflict}\}$ is the exploration mode and $\mathbf{a}_r \subseteq \mathcal{I}$ is a set of anchor insights, with $\mathcal{I}$ denoting the insight nodes in the current graph. The modes extend the frontier (\texttt{local}), deepen a finding (\texttt{refine}), resolve contradictions (\texttt{conflict}), or explore freely from the schema alone (\texttt{global}) (details in Appendix~\ref{app:rationale-modes}).

\paragraph{Generator.}
The Generator receives the discovery goal, contextual insights from the graph, a summary of over-used and under-explored variables, and previously rejected hypotheses to avoid repetition. It conducts a two-phase flow in which it first selects a thematic direction from a concept menu, then refines it into concrete testable hypotheses using variable-level metadata retrieved via RAG:
\vspace{-0.5ex}
\begin{align*}
A_{\text{gen}} : (m_r,\; \mathcal{C}_r,\; g,\; \mathcal{D}\;;\; \mathcal{F},\; \mathcal{J}) \;\longrightarrow\; \mathcal{H}_r,
\end{align*}

\vspace{-0.75ex}
\noindent where $\mathcal{C}_r$ is the contextual insight set retrieved by the Historian, $\mathcal{F}$ is the refinement queue, $\mathcal{J}$ is the rejection store, and $\mathcal{H}_r = \{h_1, \ldots, h_B\}$ is a batch of $B$ candidate hypotheses.

\paragraph{Coder/Runner.}
Each surviving hypothesis is translated into executable Python code implementing the appropriate statistical test. The agent executes the code against $\mathcal{D}$ and extracts structured results. If execution fails, the agent re-prompts with error information and retries up to $k$ attempts.
\vspace{-0.5ex}
\begin{align*}
A_{\text{code}} : (h,\; \mathcal{D}) \;\longrightarrow\; \rho_h,
\end{align*}

\vspace{-0.75ex}
\noindent where $\rho_h = (\hat{\theta}_h,\; p_h,\; \mathrm{CI}_h)$ contains the effect size, p-value, and confidence interval, or $\rho_h = \bot$ if execution fails after $k$ retries.

\paragraph{Evaluator.}
The Evaluator assesses the statistical evidence for each executed hypothesis and converts the experimental result into a validity score used by the Critic:
\vspace{-0.5ex}
\begin{align*}
A_{\text{eval}} : (\rho_h,\; \mathcal{D}) \;\longrightarrow\; s_v(h),
\end{align*}

\vspace{-0.75ex}
\noindent where $s_v(h) \in [0,1]$ is a composite validity score based on the statistical evidence for $h$.

\paragraph{Critic.}
The Critic operates at two stages of each round. Before execution, it filters the candidate batch:
\vspace{-0.5ex}
\begin{align*}
A_{\text{critic}}^{\,\text{filter}} : (\mathcal{H}_r,\; \mathcal{G}) \;\longrightarrow\; \mathcal{H}_r^{+},
\end{align*}

\vspace{-0.75ex}
\noindent where $\mathcal{H}_r^{+} \subseteq \mathcal{H}_r$ is the set of surviving hypotheses; rejected candidates are logged to $\mathcal{J}$ with their rejection reasons. After execution and evaluation, the Critic judges each result:
\vspace{-0.5ex}
\begin{align*}
A_{\text{critic}}^{\,\text{judge}} : (h,\; \rho_h,\; s_v(h),\; \mathcal{G}) \;\longrightarrow\; (s_n(h),\; d),
\end{align*}

\vspace{-0.75ex}
\noindent where $s_n(h) \in [0,1]$ is a novelty score blending structural, semantic, and epistemic surprise components, and $d \in \{\textsc{Accept},\, \textsc{Refine},\, \textsc{Reject}\}$ is the disposition decision. The pre-execution filter removes redundant, ill-formed, or previously attempted hypotheses before they consume compute. The post-execution judge prevents near-duplicate or low-novelty findings from entering the graph, and routes borderline candidates to the refinement queue $\mathcal{F}$ for future rounds.

\paragraph{Historian.}
This agent manages the insight graph through two operations:
retrieving planning context,
\vspace{-0.5ex}
\begin{align*}
A_{\text{hist}}^{\,\text{retrieve}} : (\mathbf{a}_r,\; \mathcal{G}) \;\longrightarrow\; \mathcal{C}_r,
\end{align*}

\vspace{-0.75ex}
\noindent where $\mathcal{C}_r$ is a contextual insight set assembled from the anchor nodes $\mathbf{a}_r$ selected by the Orchestrator. It also inserts accepted findings and classifies their relations:
\vspace{-0.5ex}
\begin{align*}
A_{\text{hist}}^{\,\text{insert}} : (h,\; \rho_h,\; s_v(h),\; s_n(h),\; \mathcal{G}) \;\longrightarrow\; \mathcal{G}',
\end{align*}

\vspace{-0.75ex}
\noindent where $\mathcal{G}' = \mathcal{G} \cup \{\mathrm{node}(h, \rho_h, s_v(h), s_n(h))\} \cup \mathcal{E}_h$ and $\mathcal{E}_h$ is the set of typed edges produced by relation classification.

\begin{table}[t]
\centering
\caption{Agent-level hyperparameters for reported runs.
W = WVS, A = Amazon, S = SciSciNet.}
\label{tab:agent-hparams-app}
\footnotesize
\setlength{\tabcolsep}{3pt}
\renewcommand{\arraystretch}{0.92}
\begin{tabular}{p{0.21\columnwidth}p{0.72\columnwidth}}
\toprule
\textbf{Module} & \textbf{Hyperparameters} \\
\midrule
Generator
& batch size $B=10$; context anchors $=5$ \\

Evaluator
& $p_{\min}=0.05$; effect floor $e_{\min}=0.03$; replication split $=70/30$ \\

Critic
& insight threshold $\tau_s=0.55$; structural Jaccard $=0.80$;
semantic threshold $=0.88$ \\

Coder/Runner
& timeout $=30$\,s; max retries $=3$ \\

Historian
& duplicate cosine threshold $=0.92$; retrieval order D$>$N$>$E$>$C \\

Orchestrator
& mode weights (exp./global/ref./conf.) $=0.30/0.15/0.40/0.30$;
$g_{\min}=12$; max rounds $=2{,}000$ \\

Catalog RAG (W,S)
& cooldown window $=6$, threshold $=3$ (W)/$1$ (S);
over-use cap $=15\%$; forced exploration (W) every 3 rounds \\
\bottomrule
\end{tabular}

\end{table}

\section{Extraction Pipeline: Worked Example}
\label{app:extraction-examples}

\begin{figure}[t]
    \centering
    \includegraphics[width=0.9\columnwidth]{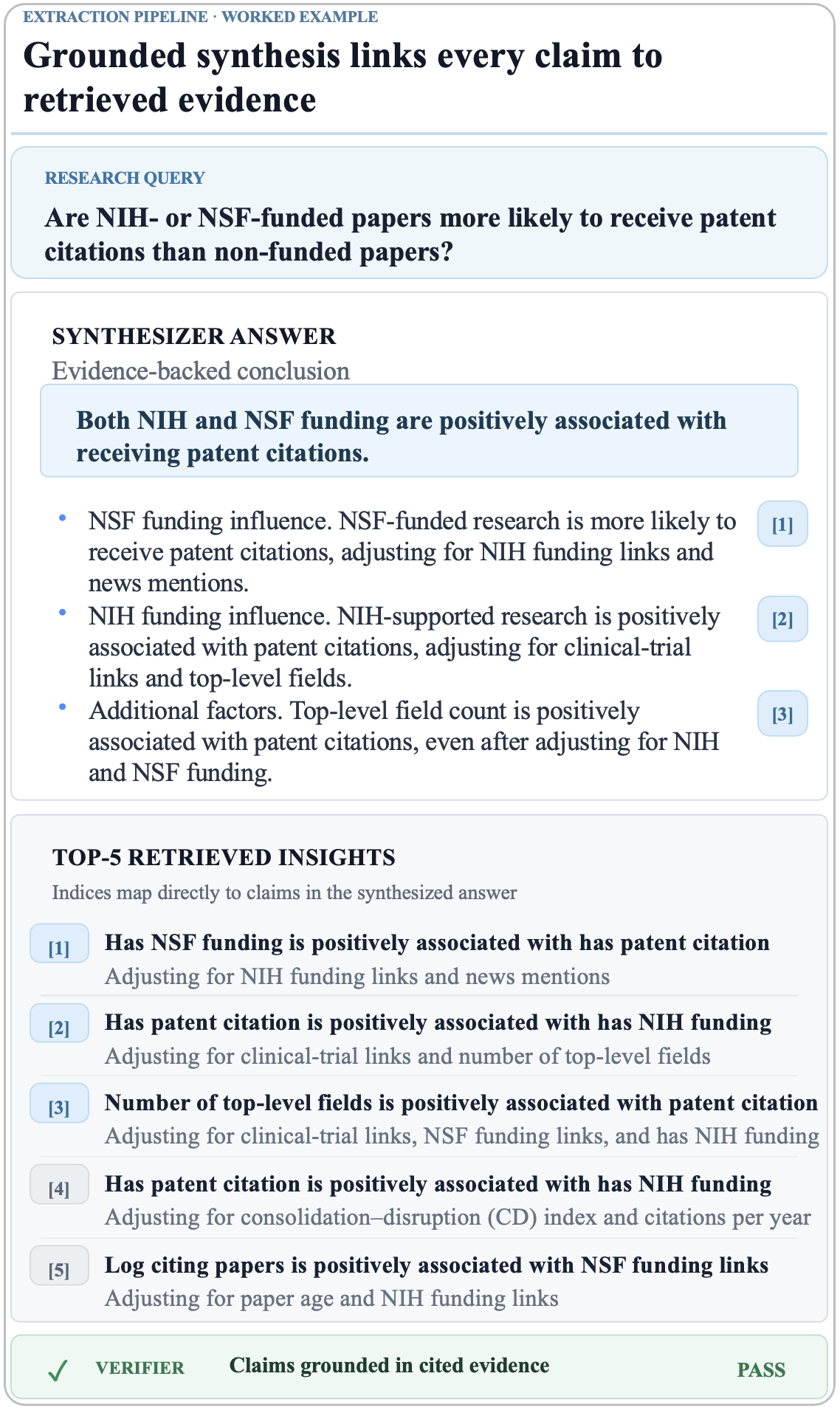}
    \vspace{-1ex}
    \caption{Worked example of the literature-grounded extraction pipeline on SciSciNet. 
    }
    \label{fig:sciscinet-extraction-example}
    \vspace{-2.75ex}
\end{figure}

Figure~\ref{fig:sciscinet-extraction-example} illustrates the extraction pipeline
on a SciSciNet query, showing the query, grounded answer, top-5 retrieved
insights, and verifier result; indices [1]--[5] refer to the retrieved insights.

\section{Human Comparison with Published \\Gold Claims}
\label{app:human-comparison}
 
The 30 insight pairs (10 per dataset) are presented through a Google Form in a fixed order: grouped by dataset (WVS, Amazon Books, SciSciNet), randomly shuffled within each dataset group, and with the A/B side assignment randomized per pair so that the \sysname insight appears as ``Finding A'' on roughly half the pairs and as ``Finding B'' on the rest. The randomization is frozen at form creation time, so all raters see an identical presentation; the assignment key is retained separately for post-hoc decoding of responses. Neither the section headings nor the findings themselves indicate which side is system-generated.

We show one representative comparison pair from each dataset in
Table~\ref{tab:case-examples}. The complete set of 30
pairs used in the human evaluation (10 per dataset) is provided in
the supplementary material.

\section{Hypothesis Space Enumeration}
\label{app:hypothesis-space}

To contextualize the scale of the discovery task, we enumerate the number of distinct hypotheses expressible under our hypothesis templates for each dataset. \sysname currently supports four base templates: \emph{association} ($X \to Y$), \emph{group difference} ($Y$ by $X$), \emph{interaction} ($X \times M \to Y$), and \emph{heterogeneity} ($X \to Y$ moderated by $M$). Each hypothesis is specified by (template, variable slots, controls), with variable slots filled from the dataset's testable variable pool of size $N$ and $k \geq 0$ controls drawn from the remaining pool.

\paragraph{Counting.} The two pair-based templates each contribute $N(N-1)$ ordered variable assignments, and the two triple-based templates each contribute $N(N-1)(N-2)$. Adding $k$ controls multiplies pair counts by $\binom{N-2}{k}$ and triple counts by $\binom{N-3}{k}$. Summing across templates gives Table~\ref{tab:hyp-total}. These are lower bounds; richer hypothesis templates would expand the space further.

\begin{table}[h]
\centering
\small
\setlength{\tabcolsep}{4pt}
\renewcommand{\arraystretch}{0.95}
\caption{Total hypothesis counts across all four templates as a function of the number of controls $k$. $N$ is the number of testable variables in each dataset. 
}
\label{tab:hyp-total}
\vspace{-1.5ex}
\begin{tabular}{crrr}
\toprule
$k$ &
\shortstack{Amazon\\($N=22$)} &
\shortstack{WVS\\($N=198$)} &
\shortstack{SciSciNet\\($N=87$)} \\
\midrule
0 & $1.94{\times}10^{4}$ & $1.54{\times}10^{7}$  & $1.29{\times}10^{6}$ \\
1 & $3.70{\times}10^{5}$ & $3.00{\times}10^{9}$  & $1.08{\times}10^{8}$ \\
2 & $3.34{\times}10^{6}$ & $2.91{\times}10^{11}$ & $4.49{\times}10^{9}$ \\
3 & $1.90{\times}10^{7}$ & $1.87{\times}10^{13}$ & $1.23{\times}10^{11}$ \\
4 & $7.61{\times}10^{7}$ & $8.98{\times}10^{14}$ & $2.48{\times}10^{12}$ \\
5 & $2.29{\times}10^{8}$ & $3.43{\times}10^{16}$ & $3.98{\times}10^{13}$ \\
\bottomrule
\end{tabular}
\vspace{-9pt}
\end{table}

\paragraph{Implications.} Even at $k=0$ the space reaches $10^7$ for WVS, and a modest five controls pushes it past $10^{16}$, which is well beyond any feasible testing budget. Exhaustive enumeration is therefore not a viable baseline, and the bottleneck shifts from hypothesis \emph{generation} to hypothesis \emph{selection}: deciding which candidates are worth executing given what has already been learned. \sysname's insight graph and Critic agent target exactly this selection problem.

\section{Orchestration and Insight-Graph Relations}
\label{app:rationale}

We detail the Orchestrator modes and Historian edge types.

\subsection{Orchestrator modes}
\label{app:rationale-modes}
The Orchestrator samples one of four modes per round, each with its
own anchor-selection strategy.
\texttt{exploration} extends the frontier by anchoring on
module-boundary, depth-frontier, or under-explored nodes.
\texttt{refinement} scores insights by a blend of overall and
validity score plus a bonus for promotable \texttt{assoc}/\texttt{interact}
nodes.
\texttt{conflict\_resolution} scans for insights estimating the same
relationship but with opposite effect directions; the pair becomes the
round's anchors, or the round falls back to a random anchor if none is
found. \texttt{global\_exploration} sends no anchors at all, allowing the
Generator to ideate without local graph-conditioned context.

Mode weights start from a fixed base and are adjusted each round based
on graph state --- insight count, cluster structure, chain density,
and conflict presence --- rather than round number. This lets the same
policy handle datasets that saturate quickly and datasets that run
long without any schedule tuning.

\subsection{Historian edge types}
\label{app:rationale-edges}

Accepted insights are linked to the anchors and context insights supplied to the round, while a separate structural pass can additionally connect them to lower-depth parent insights elsewhere in the graph. For each linked pair, the Historian's classifier assigns one of four relation types by a deterministic rule over metadata. \textsc{Contradicts} captures directional disagreement on the same estimand. \textsc{Deepens} captures progression along the depth chain with the previous insight's core variables contained in the new one. \textsc{Extends} captures cross-module bridging when the two insights live in disjoint thematic modules. \textsc{Narrows} is the fallback when none of the preceding relation rules applies.

\section*{Ethical Considerations}

Autonomous knowledge discovery introduces risks beyond those of conventional data analysis. Because \sysname explores many hypotheses with limited human intervention, it may surface spurious, misleading, or biased associations; moreover, errors accepted into the persistent insight graph may influence subsequent rounds and compound over time. The system also relies on LLM-generated analysis code, which may contain implementation errors or inappropriate statistical choices despite successful execution. We therefore treat generated insights as exploratory findings rather than established or causal conclusions, and recommend independent verification, replication, and expert review before downstream use. Additional care is needed for datasets containing demographic, political, religious, or other sensitive attributes, where automatically generated associations could reinforce stereotypes or be interpreted without sufficient context. LLM-based hypothesis generation may further bias which questions and populations receive attention. For deployment on private or identifiable data, appropriate de-identification, access controls, and restrictions on persistent storage are necessary. \sysname is intended to support scientific exploration, not autonomous high-stakes decision making.

\bibliographystyle{ACM-Reference-Format}
\bibliography{sample-base}

\clearpage

\section*{Supplementary Material}

\subsection*{S1. Statistical Estimators for Deterministic Templates}
\label{app:dsl-estimators}

For the reported experiments, hypotheses expressed using the four supported
templates---\texttt{assoc}, \texttt{diff}, \texttt{interact}, and
\texttt{heterogeneity}---are parsed into a typed specification and compiled
through a fixed statistical pipeline without an LLM call. Variables and controls
must resolve to columns in the dataset; otherwise the hypothesis is rejected
rather than executed with a modified specification. The estimator is selected
deterministically from the template, outcome type, presence of controls or
weights, and categorical cardinality. Table~\ref{tab:dsl-estimators} summarizes the estimator and reported effect used by each template.

For binary outcomes, regression-based templates use logistic regression rather
than a linear model. The target log-odds coefficient is retained together with
its odds ratio and is additionally mapped to a standardized effect,
$d=\log(\mathrm{OR})\sqrt{3}/\pi$, for consistent downstream scoring.
For multi-level categorical terms, the reported omnibus effect is necessarily
unsigned; directional interpretation is therefore restricted to individual
contrasts.

\paragraph{Preprocessing and variable types.}
Only variables required by the specification are retained. Dataset-specific
missing-value codes are first recoded as missing, followed by complete-case
analysis over variables entering the test; the retained-data fraction is recorded
as coverage. No standardization, winsorization, logarithmic transformation, or
other data-dependent transformation is introduced by the DSL. Categorical
variables are represented using treatment coding. Sparse categorical levels are
pooled before fitting, and high-cardinality specifications that remain unsuitable
for dummy expansion are rejected. Ordinal variables retain their recorded numeric
ordering in association and regression models; for an unadjusted two-group
difference, outcomes with fewer than seven observed levels are instead routed to
the rank-based Mann--Whitney test.

\paragraph{Inference and uncertainty.}
Test statistics and $p$-values come directly from the corresponding statistical
procedure: Pearson, Welch, Mann--Whitney, or Kruskal--Wallis tests for their
non-regression cases; coefficient tests or Type-II $F$-tests for OLS/WLS models;
and Wald tests for logistic regression. Regression inference uses the fitted
model's model-based standard errors. Raw model coefficients and their confidence
intervals are retained in the result record. For standardized summary effects,
95\% intervals are obtained using Fisher's $z$ transformation for correlation-type
effects, standard effect-size approximations for Cohen's $d$, delta-method
approximations for omnibus $\eta^2/\epsilon^2$, and the transformed Wald interval
for logistic models. When a weight is explicitly specified, the corresponding
linear model is fit by weighted least squares.

Finally, the executor applies minimum-sample, variance, group-size, and
categorical-cardinality checks before fitting. Specifications that cannot support
the requested estimator, or fits producing invalid numerical results, return no
empirical finding rather than silently switching the hypothesis. These estimates
are used as exploratory evidence for discovery and ranking rather than as causal
estimates.

\begin{table}[t]
\centering
\small 
\setlength{\tabcolsep}{4pt}
\caption{Statistical estimators used by the deterministic hypothesis templates.}
\label{tab:dsl-estimators}
\vspace{-2ex}
\begin{tabular}{p{0.2\columnwidth}p{0.72\columnwidth}}
\toprule
\textbf{Template} & \textbf{Estimator and reported effect} \\
\midrule
\texttt{assoc}
& Pearson correlation for unadjusted numeric variables; otherwise
OLS/WLS regression. A one-degree-of-freedom term is summarized by
(partial) $r$; a multi-level categorical predictor is tested jointly
with Type-II ANOVA and summarized by partial $\eta^2$. \\

\texttt{diff}
& For two groups without controls, Welch's $t$-test with Cohen's $d$
is used for continuous outcomes, while low-cardinality ordered outcomes
use Mann--Whitney $U$ with rank-biserial $r$. More than two groups use
Kruskal--Wallis with $\epsilon^2$. With controls or weights, the
comparison is expressed as an OLS/WLS model, using a contrast for two
groups or a Type-II omnibus test for multiple groups. \\

\texttt{interact}
& Regression of the form $Y \sim X * Z + C$, where $C$ denotes optional
controls. A single interaction coefficient is summarized by partial
$r$; interactions involving a multi-level categorical variable are
tested jointly and summarized by partial $\eta^2$. \\

\texttt{heterogeneity}
& Regression of the form $Y \sim X * G + C$, testing whether the
$X$--$Y$ association varies across groups $G$. Binary groups yield a
single interaction contrast; multi-level groups use a Type-II omnibus
interaction test. \\
\bottomrule
\end{tabular}
\vspace{-1ex}
\end{table}

\subsection*{S2. Agent Hyperparameters and Prompts}
\label{app:prompts}

Here we present 
the prompt templates used by each in-loop agent
and each evaluation judge. All in-loop calls target Qwen 2.5-14B (\texttt{Q4\_K\_M}) served via 
Ollama; the four LLM judges in the open-ended evaluation use 
\texttt{claude-opus-4-7}, \texttt{Gemini-3}, \texttt{DeepSeek-V3}, 
and \texttt{gpt-5.5}. Settings below each prompt reflect the live 
configuration.
 
\subsubsection*{S2.1 Generator}
\label{app:prompts-generator}

For all three datasets, the Generator uses a two-phase flow.
Phase 1 selects a high-level thematic direction from a
dataset-specific concept menu without referring to concrete variables.
Phase 2 retrieves relevant variable metadata via TF--IDF RAG and
refines the resulting general idea into a batch of concrete,
testable hypotheses expressed using the supported DSL templates.

\vspace{1ex}
\paragraph{Phase 1: Theme Selection.}
The dataset-specific persona, goal, focus area, exploration mode, and
concept menu are inserted into the template below. Depending on the
current exploration state, the prompt may additionally include blocks
describing recently over-used themes, previously explored directions,
promising near-miss findings, or feedback from prior retrieval
mismatches.

\begin{promptbox}{Generator -- Phase 1: Theme Selection}
\small 
You are a research scientist proposing potentially novel and
scientifically meaningful research ideas for a large-scale empirical
dataset. Your goal is to identify high-level research directions that
connect conceptually distinct themes and could lead to surprising,
testable findings.

Goal: \pvar{goal}

Focus area: \pvar{focus_area}

Mode: \pvar{mode}

\textbf{CONCEPT MENU (pick up to TWO themes by number):}

\pvar{menu_text}

\pvar{avoid_block}

\pvar{explored_block}

\pvar{refinement_block}

\pvar{grounding_block}

\textbf{Rules:}

- Do NOT mention any variable codes or column names
(no Qxx, no B\_COUNTRY, etc.).

- Keep this stage high-level and thematic. Do NOT commit to a
specific statistical template yet.

- \pvar{phase1_wording_hint}

- AIM FOR SURPRISE: prefer ideas that would contradict common
assumptions, reveal unexpected connections between distant themes,
or show that an effect reverses for certain subgroups.

Return ONLY a JSON object with these fields:

\{
  \pcode{"primary_theme_id"}: <int 1..N>,

  \pcode{"secondary_theme_id"}: <int 1..N or null>,

  \pcode{"research_question"}: "<one sentence>",

  \pcode{"core_constructs"}:
      "<2--4 short noun phrases, comma-separated>",

  \pcode{"hypothesized_relations"}:
      "<1--2 short clauses in natural language>",

  \pcode{"confounders"}:
      "<3--6 items; include demographics if plausible>",

  \pcode{"anchors"}:
      ["<6--12 short codebook-like phrases for retrieval>"],

  \pcode{"surprise_angle"}:
      "<one sentence: what would be counter-intuitive>"
\}

\end{promptbox}

\noindent\textit{Settings.}
Temperature 0.6 for exploration and global exploration, and 0.45
otherwise. Up to 3 retries on JSON parsing or format failure.

\vspace{1ex}
\paragraph{Phase 2: Variable-Level Refiner.}
The Phase 1 output is flattened into key--value lines and supplied as
the general idea. TF--IDF retrieval provides the relevant variable
cards, including column names, labels, scales, variable types,
questions, and brief semantic descriptions. A recent insight summary
and variable-cooldown block are included when applicable.

\begin{promptbox}{Generator -- Phase 2: Variable-Level Refiner}
\small 
You are a research scientist translating a high-level research idea
into concrete, statistically testable hypotheses over a structured
dataset. Convert the GENERAL IDEA into hypotheses using ONLY the
provided column names.

\textbf{GENERAL IDEA:}

\pvar{idea_block}

\pvar{history_block}

\textbf{AVAILABLE VARIABLES (schema + glossary). Use ONLY exact
column names:}

\pvar{schema_with_glossary}

\textbf{Rules:}

- Output exactly \pvar{B=10} hypotheses as a numbered list.
Nothing else.

- Each hypothesis must be a single DSL line starting with exactly
one of:
\pcode{assoc(...)},
\pcode{diff(...)},
\pcode{interact(...)}, or
\pcode{heterogeneity(...)}.

- Simple associations and group differences are allowed and
encouraged when appropriate.

- ``controlling for ...'' is OPTIONAL. If included, list 1--4
plausible controls (prefer demographics) using ONLY existing
column names.

- \pvar{dataset_specific_rules}

- Do NOT append explanations after ``:''; output only DSL lines.

- Do NOT put weight columns in ``controlling for''; if a weight is
needed, add \pcode{weights=<weight_column>} at the end.

\textbf{Internal step (do NOT output):}

Map each construct/anchor phrase to 1--2 columns from the glossary,
then instantiate kernels.

\textbf{PRIORITY ORDER (most important first):}

1. CROSS-MODULE: X and Y MUST come from different thematic sections
of the dataset. Connecting distant themes is far more valuable than
testing within-section patterns.

2. NOVELTY: Prioritize hypotheses not already in the existing
knowledge base. Simple associations are fine early on; if many are
known, explore conditional effects
(\pcode{interact}, \pcode{heterogeneity}).

3. SURPRISE: At least 2 hypotheses should test a relationship most
domain experts would NOT expect.

4. DIVERSITY: Vary outcomes/predictors across the batch.

5. CORRECTNESS: All hypotheses must use valid column names and
DSL syntax.

\textbf{DSL forms}
(one per line; examples show optional controls/weights):

- \pcode{assoc(X, Y)}
[optional: controlling for C1, C2]
[\pcode{weights=W}]

- \pcode{diff(Y, by=G)}
[optional: controlling for C1, C2]
[\pcode{weights=W}]

- \pcode{interact(X * Z -> Y)}
[optional: controlling for C1, C2]
[\pcode{weights=W}]

- \pcode{heterogeneity(Y ~ X | G)}
[optional: controlling for C1, C2]
[\pcode{weights=W}]

\textbf{Forbidden examples:}

- \pcode{assoc(X, Y | G)}

- \pcode{diff(Y by G)}

- \pcode{... : explanatory sentence}

Output only the numbered hypotheses.

\pvar{variable_cooldown_block}

\end{promptbox}

\noindent\textit{Settings.}
Temperature 0.6 for exploration and 0.4 otherwise.
Up to 3 retries if the returned list size mismatches $B$.


\subsubsection*{S2.2 Critic}
\label{app:prompts-critic}

The Critic makes two LLM calls per candidate: a cheap pre-execution
novelty estimate that can short-circuit the compute cost, and a
post-execution judge that scores epistemic surprise once statistical
results are available.

\vspace{1ex}
\paragraph{Pre-Execution Novelty Estimate.}
Up to ten retrieved neighboring insights are inserted as the local
knowledge context for the candidate hypothesis.

\begin{promptbox}{Critic -- Pre-Execution Novelty Estimate}
\small 
You are acting as a scientific peer reviewer judging the
\emph{novelty} of a single hypothesis about this dataset domain:

\pvar{domain_description}

Overall goal:

\pvar{goal}

Existing related insights (may already be known):

- \pvar{insight_sentence_1}

- \pvar{insight_sentence_2}

...

Candidate hypothesis:

``\pvar{hypothesis.sentence}''

Rate ONLY its \emph{novelty} relative to the existing insights and
to common-sense expectations about such data.

Respond in strict JSON with fields:

- \pcode{"novelty"}: a number between 0.0 and 1.0

- \pcode{"reason"}: a short string explanation.

Example:

\{
  \pcode{"novelty"}: 0.72,

  \pcode{"reason"}:
  "Combines reviewer activity and product format in a non-trivial way."
\}

\end{promptbox}

\noindent\textit{Settings.}
Temperature 0.2 with a 256-token budget. The system prompt is:
\emph{``You are a scientific peer reviewer. Output valid JSON only.''}

\vspace{1ex}
\paragraph{Post-Execution Epistemic-Surprise Judge.}
After hypothesis execution, the Critic receives the empirical finding
together with the dataset domain and up to ten retrieved neighboring
insights, and evaluates its scientific coherence, novelty, and
epistemic surprise.

\begin{promptbox}{Critic -- Post-Execution Epistemic-Surprise Judge}
\small 
You are a research scientist building a knowledge base of empirical
findings from a dataset.

Your task is to evaluate whether a newly discovered pattern is
scientifically meaningful.

\textbf{INPUT DATA}

1. \textbf{The Hypothesis:}
``\pvar{display_sentence}''

2. \textbf{Dataset Domain:}
\pvar{domain_description}

3. \textbf{Existing Knowledge:}

- \pvar{insight_sentence_1}

- \pvar{insight_sentence_2}

...

\textbf{YOUR TASK}

First, ask yourself: \textbf{what mechanism could explain this
relationship?}

If you cannot articulate a plausible causal or theoretical pathway
in 1--2 sentences, the finding is INCOHERENT regardless of
statistical significance.

Then judge how novel and valuable the finding is.

\textbf{SCORING RUBRIC (0--10)}

\textbf{0--2: TAUTOLOGICAL}

- Variables measure the same underlying construct, or the
relationship is a coding artifact, definition, or mathematical
identity.

- Classification: TRIVIAL. Action: REJECT.

\textbf{3--4: INCOHERENT}

- No mechanism explaining why these variables would be related.

- Combination is arbitrary; no researcher would hypothesize this.

- Classification: TRIVIAL. Action: REJECT.

\textbf{5--6: FOUNDATIONAL}

- Clear mechanism; direction a domain researcher would predict.

- Quantifies a known phenomenon; confirming it across contexts is
valuable.

- Classification: FOUNDATIONAL. Action: ACCEPT.

\textbf{7--8: NOVEL}

- Plausible mechanism; relationship not commonly tested.

- Bridges research areas, reveals a hidden moderator, or quantifies
an assumed-but-unverified effect.

- Classification: NOVEL. Action: STRONG ACCEPT.

\textbf{9--10: SURPRISING}

- Finding contradicts domain expectations or reveals a
reversal/amplification in subgroups.

- Classification: SURPRISING. Action: STRONG ACCEPT.

\textbf{RESPONSE FORMAT (JSON ONLY)}

\{
  \pcode{"mechanism"}:
  "<1--2 sentences on WHY these variables would relate,
  or 'none' if incoherent>",

  \pcode{"reasoning"}:
  "<step-by-step novelty analysis>",

  \pcode{"classification"}:
  "TRIVIAL | FOUNDATIONAL | NOVEL | SURPRISING",

  \pcode{"score"}:
  <float between 0.0 and 10.0>,

  \pcode{"critique"}:
  "<one specific suggestion to improve novelty>"
\}

\end{promptbox}

\subsection*{S3. Complete \sysname--Gold Claim Comparison Pairs}
\label{app:prompts}

Tables~\ref{tab:human-scisci}--\ref{tab:human-wvs} report all 30 pairs used in the open-ended human evaluation, with 10 pairs per dataset. Each row presents an AutoKD-discovered insight alongside its matched published gold sub-claim. As described in Appendix C, the displayed A/B order was randomized in the evaluation form to preserve blinding.

\begin{table*}[t]
\centering
\footnotesize
\caption{Complete SciSciNet CS comparison pairs between \sysname-discovered insights and matched published gold claims.}
\label{tab:human-scisci}
\vspace{-3ex}
\setlength{\tabcolsep}{4pt}
\renewcommand{\arraystretch}{1}
\begin{tabularx}{\textwidth}{c >{\raggedright\arraybackslash}X >{\raggedright\arraybackslash}X}
\toprule
\# & \sysname system insight & Gold sub-claim \\
\midrule
1 & Solo-authored papers have less extreme novelty in their most atypical reference pair than co-authored papers. & NIH- or NSF-funded papers show higher novelty (more atypical reference combinations) than non-funded papers~\cite{yang2024unveiling}. \\
2 & Heavily cited papers span fewer broad academic fields. & NIH-funded and NSF-funded papers are more likely to receive patent citations than non-funded papers~\cite{yang2024unveiling}. \\
3 & The relationship between the number of fields a paper spans and its 3-year citation count depends on whether the paper is solo-authored. & The association between female first authorship and citation counts varies by academic field~\cite{shi2024women}. \\
4 & Papers with more 5-year citations tend to have later sleeping-beauty awakening years --- the year a paper's citation count finally starts surging, often decades after publication; a later year means a longer dormancy. & The negative relationship between a first author's publication count and disruption (a measure of how much a paper displaces prior work rather than building on it; higher = more disruptive) is stronger in STEM fields than in humanities or social sciences~\cite{li2024productive}. \\
5 & Papers published longer ago earned fewer citations in their first three years after publication than more recent papers did --- early citations are measured in a fixed three-year window, so this reflects that early-career citation rates have grown over time rather than older papers losing attention. & Papers with higher disruption scores (a measure of how much a paper displaces prior work rather than building on it; higher = more disruptive) are less likely to receive patent citations; however, this largely reflects the metric's tendency to score low-citation papers higher rather than a property of disruptive research itself~\cite{yang2025disruptive}. \\
6 & Heavily cited papers have shorter sleeping-beauty burst durations than other papers --- once a dormant paper's citations start surging, the surge lasts less time before levelling off. & NIH- or NSF-funded papers do not differ meaningfully in disruption scores (a measure of how much a paper displaces prior work rather than building on it; higher = more disruptive) from non-funded papers~\cite{yang2024unveiling}. \\
7 & Papers spanning more broad academic fields accumulate more total citations. & Papers with NIH or NSF funding receive more citations than non-funded papers~\cite{yang2024unveiling}. \\
8 & Papers whose last author is more likely to be female receive fewer 10-year citations. & Papers with a female first author tend to have lower reference-conventionality scores (more novel reference combinations)~\cite{shi2024women}. \\
9 & First authors whose prior papers were cited more on average tend to publish papers spanning more fields. & Papers by more productive first authors tend to cite more recently published references~\cite{li2024productive}. \\
10 & Papers with authors from more countries tend to have more atypical reference pairs (rare combinations of cited works that few other papers share). & Papers by more productive first authors have higher reference-conventionality scores (more typical reference combinations)~\cite{li2024productive}. \\
\bottomrule
\end{tabularx}
\vspace{-0ex}
\end{table*}
 
 
\begin{table*}[t]
\centering
\footnotesize
\caption{Complete Amazon Books comparison pairs between \sysname-discovered insights and matched published gold claims.}
\label{tab:human-amazon}
\vspace{-3ex}
\setlength{\tabcolsep}{4pt}
\renewcommand{\arraystretch}{1}
\begin{tabularx}{\textwidth}{c >{\raggedright\arraybackslash}X >{\raggedright\arraybackslash}X}
\toprule
\# & \sysname system insight & Gold sub-claim \\
\midrule
1 & Longer reviews tend to appear on less popular books. & Longer reviews receive more helpful votes~\cite{choi2020empirical}. \\
2 & Less popular books tend to have reviews with richer vocabulary (a higher proportion of unique words per review). & Reviews with extreme ratings (1-star or 5-star) receive more helpful votes than moderate-rated reviews~\cite{choi2020empirical}. \\
3 & The proportion of reviews containing questions has decreased over time. & Average review length in Amazon book reviews has declined over time for both 5-star and 1-star reviews~\cite{ziser2023rant}. \\
4 & Books with more 'also bought' recommendations --- other products displayed on the book's page --- tend to have reviews with richer vocabulary (a higher proportion of unique words per review). & Reviewers who have posted many reviews receive fewer helpful votes per review than occasional reviewers do~\cite{choi2020empirical}. \\
5 & Reviews of verified purchases have different vocabulary richness (proportion of unique words per review) than non-verified reviews. & Reviews with a verified purchase badge receive more helpful votes than unverified reviews~\cite{choi2025trust}. \\
6 & Books with more 'also bought' recommendations --- other products displayed on the book's page --- tend to have shorter reviews. & A book with more total reviews is associated with fewer helpful votes per individual review~\cite{choi2020empirical}. \\
7 & Books with more 'also viewed' recommendations --- other products displayed on the book's page --- tend to have reviews with richer vocabulary (a higher proportion of unique words per review). & Reviews containing user-provided images receive more helpful votes than reviews without images~\cite{choi2025trust}. \\
8 & Average book ratings have increased over time. & A book's higher average rating is associated with fewer helpful votes per review~\cite{choi2020empirical}. \\
9 & Books with a listed description tend to be more popular than books without descriptions. & The gap between a review's rating and the book's average rating is negatively associated with helpful votes~\cite{choi2020empirical}. \\
10 & More popular books tend to have more 'also bought' recommendations --- other products displayed on the book's page. & The combination of a verified purchase badge and user-provided images boosts helpful votes beyond what either signal alone would predict~\cite{choi2025trust}. \\
\bottomrule
\end{tabularx}
\vspace{-0ex}
\end{table*}
 
 
\begin{table*}[t]
\centering
\footnotesize
\caption{Complete WVS comparison pairs between \sysname-discovered insights and matched published gold claims.}
\label{tab:human-wvs}
\vspace{-3ex}
\setlength{\tabcolsep}{4pt}
\renewcommand{\arraystretch}{1}
\begin{tabularx}{\textwidth}{c >{\raggedright\arraybackslash}X >{\raggedright\arraybackslash}X}
\toprule
\# & \sysname system insight & Gold sub-claim \\
\midrule
1 & Higher social trust is negatively associated with discomfort with women earning more than their husbands. & Being in a relationship is positively associated with higher life satisfaction~\cite{itani2024statistical}. \\
2 & Higher perceived corruption is positively associated with placing greater value on obedience as a child quality. & Males report higher justification of parents beating children than females~\cite{begum2023justification}. \\
3 & More frequent religious service attendance is positively associated with stronger endorsement of nativist hiring priority when jobs are scarce. & Religious affiliation is positively associated with higher life satisfaction~\cite{itani2024statistical}. \\
4 & Stronger endorsement of male hiring priority when jobs are scarce is positively associated with greater confidence in churches. & Gender is not a significant predictor of life satisfaction~\cite{itani2024statistical}. \\
5 & Higher education level is positively associated with greater social trust. & Higher financial satisfaction is positively associated with higher life satisfaction~\cite{ng2026national}. \\
6 & Viewing immigrants' impact more positively is negatively associated with placing importance on God in one's life. & Higher education level is positively associated with lower justification of parents beating children~\cite{begum2023justification}. \\
7 & Greater confidence in television news is positively associated with feeling secure in one's neighborhood. & Higher financial satisfaction is positively associated with greater happiness~\cite{ng2026national}. \\
8 & Viewing competition as harmful rather than beneficial is positively associated with greater justification of tax cheating. & Younger respondents report higher justification of parents beating children than older respondents~\cite{begum2023justification}. \\
9 & Greater worry about a terrorist attack is negatively associated with feeling secure in one's neighborhood. & Going without basic needs (food, medicine, or housing) is negatively associated with self-rated health~\cite{basant2025resource}. \\
10 & Greater medical insecurity (having gone without needed medicine) is positively associated with stronger endorsement of male hiring priority when jobs are scarce. & Going without basic needs (food, medicine, or housing) is negatively associated with financial satisfaction~\cite{basant2025resource}. \\
\bottomrule
\end{tabularx}
\end{table*}

\end{document}